\documentclass[letterpaper, 10 pt, conference]{ieeeconf}  
\IEEEoverridecommandlockouts                               
\usepackage{graphicx}                       
\usepackage{graphics}                       
\usepackage{epsfig}                         
\usepackage[tight,footnotesize]{subfigure}  

\usepackage{amssymb,amsmath}
\usepackage{mdwmath}
\usepackage{mdwtab}
\usepackage{commath}                        
\usepackage{eqparbox}
\usepackage{amsmath,amsfonts,amssymb}

\newcommand{\beq}{\begin{equation}}
\newcommand{\eeq}{\end{equation}}
\newcommand{\bear}{\begin{eqnarray}}
\newcommand{\bears}{\begin{eqnarray*}}
\newcommand{\eear}{\end{eqnarray}}
\newcommand{\eears}{\end{eqnarray*}}
\newcommand{\bdm}{\begin{displaymath}}
\newcommand{\edm}{\end{displaymath}}
\newcommand{\lba}{\left[\begin{array}}
\newcommand{\ear}{\end{array}\right]}

\usepackage{stfloats}                       

\usepackage[T1]{fontenc}

\newcommand{\squeezeup}{\vspace{-2.5mm}} 
\title{\LARGE \bf Robot Introspection via Wrench-based Action Grammars.}
\author{Juan Rojas, Zhengjie Huang, Shuangqi Luo, Yunlong Du
Wenwei Kuang, Dingqiao Zhu, and Kensuke Harada\\
\thanks{Juan Rojas is with the School of Electromechanical Engineering in the Guangdong University of Technology in Guangzhou, China.}%
\thanks{WenWei Kuang and Ding Qiao are with the School of Software at Sun Yat Sen University in Guangzhou, China.}%
\thanks{Kensuke Harada is with the Intelligent Sys. Research Institute at AIST in Tsukuba, Ibaraki, Japan.}%
}
\begin{document}
\maketitle
\thispagestyle{empty}
\pagestyle{empty}
\bstctlcite{IEEEexample:BSTcontrol} 
\begin{abstract}
Robotic failure is all too common in unstructured robot tasks. Despite well designed controllers, robots often fail due to unexpected events.
How do robots measure unexpected events? Many do not. Most robots are driven by the sense-plan-act paradigm, however more recently robots are working with a sense-plan-act-verify paradigm.
In this work we present a principled methodology to bootstrap robot introspection for contact tasks. In effect, we are trying to answer the question, what did the robot do?
To this end, we hypothesize that all noisy wrench data inherently contains patterns that can be effectively represented by a vocabulary. The vocabulary is generated by meaningfully segmenting the data and then encoding it. When the wrench information represents a sequence of sub-tasks, we can think of the vocabulary forming sets of words or sentences, such that each subtask is uniquely represented by a word set. Such sets can be classified using statistical or machine learning techniques. We use SVMs and Mondrian Forests to classify contacts tasks both in simulation and in real robots for one and dual arm scenarios showing the general robustness of the approach.
The contribution of our work is the presentation of a simple but generalizable semantic scheme that enables a robot to understand its high level state. This verification mechanism can provide feedback for high-level planners or reasoning systems that use semantic descriptors as well.
The code, data, and other supporting documentation can be found at: http://www.juanrojas.net/2017icra\_wrench\_introspection.
\end{abstract}
\section{INTRODUCTION}\label{sec:Intro}
%
%
In autonomous scenarios, robotic failure is an undesirably frequent event. Despite well designed optimal controllers, robots often fail due to unexpected events occurring within their workspace. One aspect that leads to failure is that robots lack a sense of awareness upon executing their actions. Appropriately designed controllers give robots an action potential to reach set-points and reject disturbances; however, controllers are unable to make sense of unexpected events. From a different vantage point, a long-held paradigm in robotics has been the: ``sense-plan-act'' paradigm, but we hold that such paradigm is limited in its ability to cope with uncertainty. We would add a 4th element named such that the paradigm stood as: ``sense-plan-act-verify''.
\begin{figure}[th!]
    \centering
        \includegraphics[scale=0.85]{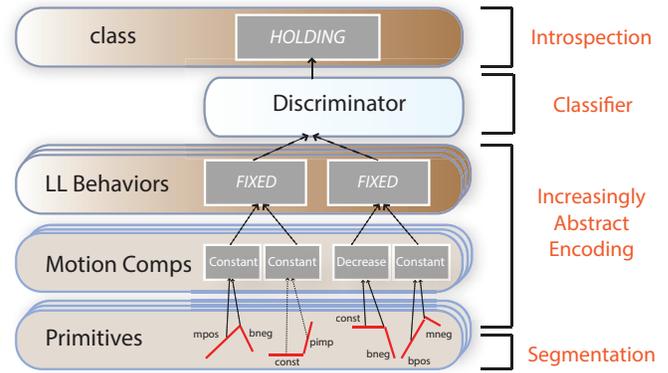}
        \caption{Our approach takes noisy wrench signals and continuously segments, encodes, and classifies to provide the robot with introspection about what it is doing.}
        \label{fig:Taxonomy}
\end{figure}
In contact tasks, wrench signal interpretation is not as straight forward as position data. The goal of the verification step is then to understand the signal evolution to help the robot understand what is doing. In doing so, it can use this information to reason about its next move. Much work in the manipulation literature has gone into identification of robot skills that are flexible and reusable \cite{2015ICRA-Kroemer-TwrdsLearnHierSkillsMultiPhaseManip,2015IJRR-Niekum-LrnGrnddFiniteStateReprUnstrucDems}. Fewer work has been done in the verification arena. Where after a task has been commanded, how do we know if it was appropriately completed or not?
In the area of verification, most work is divided under model-based techniques and data-driven techniques. If available, models of the system or the environment can be exploited to yield state estimates, though models are not always available due to system complexity \cite{2014IROS-Lowrey-ConsistentSensrFusionContactRichBehaviors,2016ICRA-Zhou-ConvexPolyForceMtnModel}. On the other hand, data-driven approaches collect data from one or more sensors and often use probabilistic \cite{2013IROS-DiLello-BayesianContFaultDetection,1998IJRR-Hovland-HMM_ProcessMonitorAsmbly} or machine learning concepts \cite{2010CASE-Rodriguez-FailureDetAsmbly_ForceSigAnalysis, 2011IROS-Rodriguez-AbortRetry, 2014ICRA-Rojas-EarlyFC} to estimate the task's state.
Data-driven techniques can then be categorized as: (i) Discrete \emph{vs.} Continuous Event Evaluation: for discrete evaluations, contact points are evaluated as a contact sequence \cite{1998IJRR-Hovland-HMM_ProcessMonitorAsmbly, 2010CASE-Rodriguez-FailureDetAsmbly_ForceSigAnalysis, 2011IROS-Rodriguez-AbortRetry, 2015ICRA-Golz-TactileSensingLearnContactKnowledge}, whilst for the continuous scenario, it is the signal evolution that is modelled \cite{2013IROS-DiLello-BayesianContFaultDetection, 2013IJMA-Rojas-TwrdsSnapSensing} and (ii) Low-Level State Estimation \emph{vs.} High-Level State Estimation: the output of the state estimation technique for event modeling can be numeric \cite{2014IROS-Lowrey-ConsistentSensrFusionContactRichBehaviors,2016ICRA-Zhou-ConvexPolyForceMtnModel, 2015ICRA-Golz-TactileSensingLearnContactKnowledge, 2011IROS-Rodriguez-AbortRetry},\cite{2010CASE-Rodriguez-FailureDetAsmbly_ForceSigAnalysis, 2007Tro-Meeussen-CtctStSegPartFilt} or semantic \cite{2013IJMA-Rojas-TwrdsSnapSensing}. The notion of a semantic representation for tasks has been used some in medical robotics by using pose information from surgical robots to measure the skill with which a surgeon performs a surgery \cite{2013MICCAI-Ahmidi-StringMotifDescrToolMotion_SkillGestures}.

This work's contribution is a principled methodology to bootstrap robot introspection for contact tasks through a continuous, data-driven, high level semantic state approach. In effect, noisy wrench signals are segmented, encoded, and classified to yield a belief about the robot's state allowing it to verify its own actions. Fig. \ref{fig:Taxonomy}, presents an overview of our approach. Our current work differs from our previous efforts by establishing formal supervised classification methods for a data-driven approach vs previously hand-picked features. Also we demonstrate the robustness of the approach by testing it with simulated and real robots in one and two arm scenarios. Our previous efforts had only tested with one arm simulations.

We hypothesize that all noisy wrench data inherently contains patterns that can be effectively represented by a vocabulary. The vocabulary is generated by meaningfully segmenting the data and then encoding it. Segmentation and encoding are performed through the Relative Change-Based Hierarchical Taxonomy (RCBHT) \cite{2013IJMA-Rojas-TwrdsSnapSensing}, the latter tries to capture relative change in wrench data by fitting straight line regression segments to the signal. From there the data is encoded into a series of increasingly abstract layers through a small set of categories yielding an action grammar for the task. We assume tasks are composed of a sequence of sub-tasks or phases \cite{2015ICRA-Kroemer-TwrdsLearnHierSkillsMultiPhaseManip}. For each sub-task the action grammar forms a set of words or a sentence that uniquely describes that phase. We study classification of sentences using two different classifiers: SVMs and Mondrian Forests. We compare their efficacy in discriminating the robot behavior three contacts tasks: (i) a one arm assembly task in simulation, (ii) the same task with the real robot, and (ii) the same assembly task but in a dual-arm scenario.

Our experimental results that the robot is able to perform state introspection with mean accuracy rates of above $95$ for both classifiers for one arm real robot trials and two are simulated trials, and slightly lower accuracies for one arm simulated trials.
The advantage of this verification system is that its semantic nature can suitably provide feedback to high-level planners or reasoning systems \cite{2015IJCAI-Konidaris_Lozano-SymbolAcquisitionForProgHLPlanning, 2013ExpBots-Matuszek-LearningParseNatlLangToCommands}.

The rest of the paper is organized as follows: Sec. \ref{sec:RCBHT} presents the segmentation and encoding steps to wrench signals, Sec. \ref{sec:Classification} introduces both classification algorithms, Sec. \ref{sec:Experiments} introduces the contact task experiment and results; Sec. \ref{sec:Discussion} discusses originality, strengths, weaknesses, and future work, and Sec. \ref{sec:Conclusion} summarizes key findings.
\section{The Relative Change-Based Hierarchical Taxonomy}\label{sec:RCBHT}
The use of the RCBHT is assumed to take place in the context of a robot task which is composed of sub-tasks otherwise known as states or phases. The segmentation of sub-tasks can take place either through any format: programming by demonstration, a finite state machine (FSM), or any other means. It is assumed that the state segmentation is known. The RCBHT \cite{2013IJMA-Rojas-TwrdsSnapSensing} enables semantic encoding of low-level wrench data. The taxonomy is built on the premise that low-level relative-change patterns can be classified through a small set of categoric labels in an increasingly abstract manner.
The RCBHT is a multi-layer behavior aggregating scheme. It is composed of three bottom-to-top increasingly abstract layers and two top-to-bottom layers. From the bottom we have the Primitive's Layer, the Motion Composition (MC) layer, the Low-Level Behavior Layer (LLB). From the top we have a verification layer and a classification layer. The taxonomy is illustrated in Fig. \ref{fig:Taxonomy}.

To bootstrap the introspection process, each of the six Force-Torque axes are separated and operated on individually. The Primitive layer partitions data for each of the six axis into linear segments that roughly approximate the original signature. For each segment, a number of  features are extracted and a gradient classification label provided. The next layer examines the gradient labels of primitive ordered-pairs. According to a gradient pattern classification criteria, the ordered pairs are categorised into a higher abstraction set--the MCs. The third layer applies the same logic to motion compositions to produce another higher abstraction--the LLBs. The advantage of having multiple layers of increasing abstraction is two-fold: (i) increasingly abstract layers yield increasingly more intuitive semantic representations of the robot behaviors thus increasing their suitability for higher-level planning and reasoning processes; (ii) the size of the dimensional space decreases by a factor of $2^x$ where $x$ is each new layer thus decreasing the computational cost and increasing speed; and (iii) noise in the system is increasingly filtered with each new layer. Having said so, one is also able to use multiple layers to learn more about the process. The top-down approach uses the classification layer to discriminate which action grammar sentences best encode the sub-tasks a robot performs.

\subsection{The Primitive's Layer}\label{subsec:Primitives}
\begin{figure}[ht]
    \centering
        \includegraphics[scale=1]{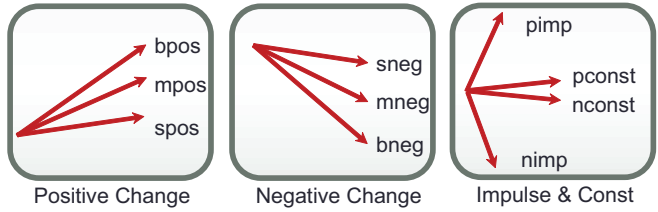}
        \caption{Gradient Classifications for Wrench Data in the Primitives Layer. There are 3 main groups: positive, negative, and const value gradients. To understand relative changes of magnitude, positive and negative gradients are divided into 4 regions: small, medium, large, and impulses.}
        \label{fig:Primitives}
\end{figure}

The Primitives layer partitions wrench data into linear data segments and classifies them according to gradient magnitude. Linear regression along with a correlation measure (the determination coefficient $R^2$) are used to segment data when a minimum correlation threshold is flagged.
Gradient classification for wrench data is fundamentally the set of three gradient value groups: positive, negative, and constant gradients. This simple classification is enough to capture relative change. Further, to understand relative magnitudes of change, positive and negative groups are subdivided into four ranges: small, medium, large, and impulses (very large). Contact phenomena is characterized by abrupt changes in wrench signals almost approximating an (positive or negative) impulse, it is such impulses that are characteristic of surface contacts.  Negative and positive gradient ranges are labelled as: ``sneg, mneg, bneg, nimp" and ``spos, mpos, bpos, and pimp" respectively. Constant gradients are those whose change is trivial. They are classified as such if their gradient magnitude is lower than the absolute value of a calibrated threshold and are labeled as ``const". A visualization of these gradient classifications can be seen in Fig. \ref{fig:Primitives}.

To generalize gradient thresholds, a calibration routine uses contextual information to determine both the largest and the near-constant gradient values in a task, which are then used to set corresponding upper and lower boundary values (see \cite{2012ROBIO-Rojas-GradientOptimization} for more details). Furthermore, seven data features are extracted from each primitive segment: average value, maximum and minimum value, the starting and ending time, the gradient value, and the corresponding label. Such features can be used freely for different approaches. These, for instance, have been used in probabilistic methods for early failure detection \cite{2014ICRA-Rojas-EarlyFC,2014Humanoids-Rojas-ContextualizedEArlyFailureCharac} and Bayesian filtering \cite{2012Humanoids-Rojas-pRCBHT}.
\subsection{The Motion Composition's Layer}\label{subsec:MCs}
\begin{figure}[ht]
    \centering
        \includegraphics[scale=1]{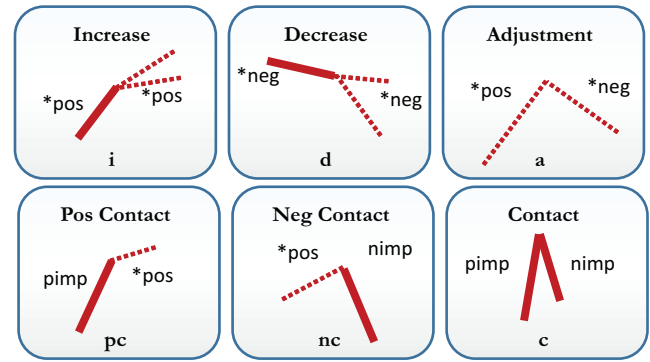}
        \caption{Illustration of possible generation of MCs. MCs are constructed by ordered-pairs of primitives. Six main groupings are shown. Each MC appears with its name, corresponding label, and ordered primitive pair. Dotted lines represent segments that can take primitives of different magnitudes, i.e. small-, medium-, or big-positive ones}.
        \label{fig:MotComps}
\end{figure}

The MC layer classifies ordered-pairs of primitives into seven categories: Adjustment, Increase, Decrease, Constant, Contact, and unstable motions. These 7 categories still represent the positive, negative, and near-constant representation of the previous layer but also give rise to adjustments and unstable motions. Adjustments are fundamentally primitive ordered-pairs that have a positive-negative or negative-positive (of the varying ranges) transitions. Adjustments represent wrist motions in which a quick ``back-and-forth" jerk action is seen. They are typical in alignment and insertion operations where force controllers minimize residual errors. Positive and negative gradients are grouped in this way to maximize the likelihood of grouping any such jerks regardless of slight variations in magnitude. For Increase, Decrease, and Constant categories they group contiguous (small-to-big) positive, (small-to-big) negative, and constant primitives respectively. For Contacts any positive or negative contact followed by any (small-to-big) negative primitive or (small-to-big) positive primitive yields a Contact, as well as a positive contact followed by a negative contact or vice-versa. These groupings along with their respective labels are illustrated in Fig. \ref{fig:MotComps}.
Eleven features were collected for each MC: composition label, average value, root means square (RMS), amplitude, first and second primitive labels, starting and ending times for both primitives, and average time for both primitives.
\subsection{The Low Level Behavior Layer}\label{subsec:LLBs}
\begin{figure}[ht]
    \centering
        \includegraphics[scale=1]{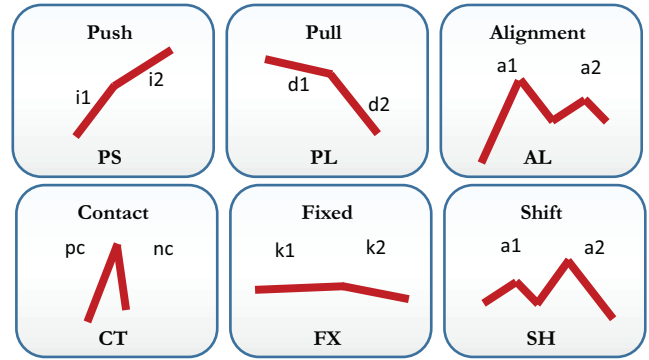}
        \caption{Illustration of possible generation of LLBs. LLBs are constructed by ordered-pairs of MCs. Six main groupings are shown. Each LLB appears with its name, corresponding label, and ordered MC pair.}
        \label{fig:LLBs}
\end{figure}

\begin{figure*}
    \centering
        \includegraphics[width=1.0\linewidth]{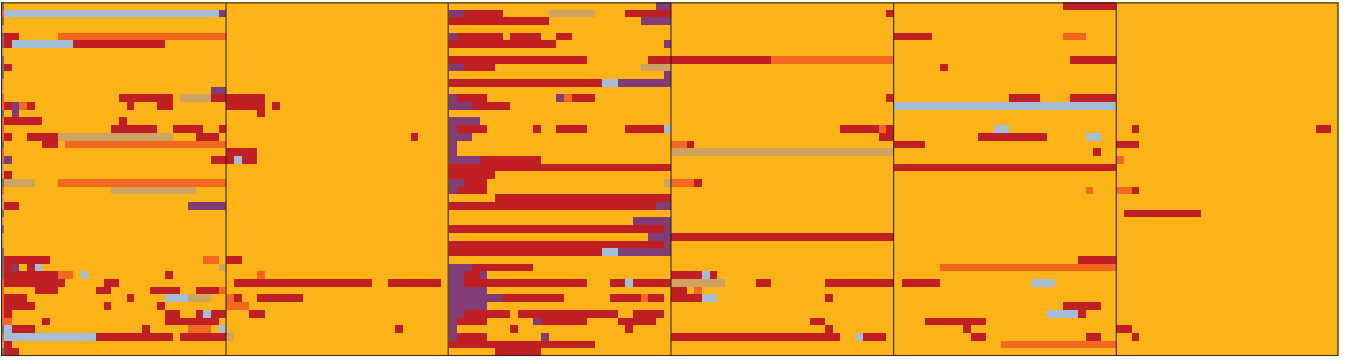}
        \caption{A color coded representation of the action grammar produced by LLBs for six FT axes stacked next to each other. For each stack, each column represents a different word. Each row represents a different trial. Patterns are evident at simple view across each FT axes. It is also visible that different axes contains very different types of information.}
        \label{fig:LLBcolorCodedMap}
\end{figure*}

As with the previous layer, the LLB layer classifies ordered-pairs of MCs into seven categories: Push, Pull, Fixed, Contact, Alignment, Shift, and Noise. The classification criteria is similar to the MC level but extends the definition of adjustments into increasingly stable adjustments (alignments) or increasingly unstable (shifts). Push and Pull group contiguous pairs of increase and decrease MCs respectively. Fixed and Contact group contiguous pairs of constant or contact MCs respectively. One major difference between the MC level and this level is the introduction of a shifting behavior `SH'. Shifts and Alignments are similar but differ in that, whenever there are two contiguous adjustment compositions, if the second composite's amplitude is larger than the first, then it is a Shift, otherwise an Alignment. In effect, Alignments are adjustments that converge while Shifts become unstable over time. These groupings along with their respective labels are illustrated in Fig. \ref{fig:LLBs}.
The same eleven features recorded in the previous layer are recorded here but with respect to MCs. There is one exception: the RMS feature was switched for a maximum value feature. A color coded representation of the action grammar produced by LLBs for each of the six FT axes can be seen in Fig. \ref{fig:LLBcolorCodedMap}.
\subsection{Abstraction and Refinement}\label{subsec:Refinement}
The three bottom-top layers have an embedded redundancy in them. Pairs of positive gradients, become `Increase' MCs, which become `Push' LLBs. We have found that this grouping paired with a filtering mechanism (described shortly) encode behaviors very well. Filtering is executed for every layer in the taxonomy and is based on three different criteria: (i) a time-duration context, (ii) repeated behaviors, and an (iii) amplitude value context. The first category eliminates time negligible behaviors. The second category merges repeated behaviors. The third category examines adjacent signals magnitudes, and if one signal is much larger than the other (with the exception of impulse signals), the lower amplitude signal is merged with the other. Each layer runs a filtering cycle 2-3 times to significantly reduce the label number and yield a most representative action grammar.  For example, an MC literal like `iiddiidd', would be simplified as follows: $`iiddiidd' \rightarrow `idid' \rightarrow `aa' \rightarrow `a'$. Actions are grouped and filtered. These in turn are further abstracted in the LLB layer, where they again undergo one more grouping and filtering round. In the end, wrench signatures can be modeled for each axis as shown in Fig. \ref{fig:LLBcolorCodedMap}. For further details about thresholds and specific filtering criteria see \cite{2013IJMA-Rojas-TwrdsSnapSensing}.
\section{Classification Mechanisms}\label{sec:Classification}
As part of the top-bottom RCBHT scheme, the classification layer enables the robot to associate the relevant portion of the action grammar in one sub-task with the sub-task. The classification mechanism assumes that the proper segmentation is taking place to appropriately sub-divide the task whether in nominal scenarios or anomalous scenarios. A number of autonomous segmentation works for manipulations tasks can be found in: \cite{2015ICRA-Kroemer-TwrdsLearnHierSkillsMultiPhaseManip}\cite{2015IJRR-Niekum-LrnGrnddFiniteStateReprUnstrucDems}. Thus, under a data-driven classification approach, the goal is to help the robot identify what phase he is actually executing given some control goal. This is particularly useful when unexpected events occur and the robot is unable to accomplish the task which he has been commanded to execute. This does necessitate that the robot trains under the different scenarios. And for cases in which a new scenario occurs it must be duly labelled. In this work we have tested classification with two discriminators, a standard SVM classifier \cite{2004SnC-Smola-SVRTutorial} presented in Sec. \ref{subsubsec:svm} and a more recent efficient online random forest algorithm named Mondrian Forests presented in Sec. \ref{subsubsec:forests} \cite{2014NIPS-lakshminarayanan-MondrianForests}. The next section provides a high level overview of the algorithms adapted to our work.
\subsection{Support Vector Machines}\label{subsubsec:svm}
Linear Support Vector Machines (SVMs) approximate a boundary to separate binary classes through a hyperplane for large feature spaces. The feature vector is used to learn a hyperplane: $\omega^T x − b=0$, where $\omega$ are the weights and $b$ is the bias from the zero point. In effect, the separation of each training point from the hyperplane is the functional margin $\hat{\gamma}^(i)$ and can be modeled as:
\begin{equation}\label{eqtn:hyperplane}
    \hat{\gamma}^{(i)} = y^{(i)}(\omega^{(i)}x + b)
\end{equation}
Here the pair ${y(i), x(i)}$ represents the class as $y^{(i)} \in {1,−1}$ and $x^{(i)}$ is the input vector
for training and testing. The SVM optimizes the functional margin by maximizing the distance to both true and
failure cases by solving the quadratic programming problem:
\begin{gather}\label{eqtn:gamma}
  \max (\gamma)                                       \nonumber       \\
  \quad \mbox{ s.t.} \quad \gamma = \min_{i=1,..,m} \hat{\gamma},
\end{gather}
where, $\gamma$ is the geometrical margin of the input points from the hyperplane. The larger the geometrical margin the more
accurate the classifier. Our linear classifier was tested with a linear, polynomial, and a radial basis function as kernels using Scikit's machine learning library \cite{scikit-learn}.
\subsection{Mondrian Forests}\label{subsubsec:forests}
Random forests have been used for classification and regression with competitive computational and predictive performance, suitable for online classification tasks. Popular offline random forest variants \cite{2001ML-Breiman-RandomForests} use batch processing for training, but until recently online variants needed more training data than offline batch training to achieve similar results. Mondrian Forests are also ensembles of randomized decision trees but the difference with traditional approaches is that they can be grown incrementally online and achieve the same distribution as that offline batch decision tree processing \cite{2014NIPS-lakshminarayanan-MondrianForests}. They achieve better predictive performance compared to existing online methods and nearly match those of state-of-the-art batch random forests, however they are more than one order of magnitude faster.

Given $N$ labeled training samples $(\mathbf{x_1},y_1),...,(\mathbf{x_N},y_N) \in \mathbb{R}^D \times \gamma $, the task is to predict labels $y \in \gamma$ for unlabeled test points $\mathbf{x} \in \mathbb{R}^D$. In fact, for our work, each label $x_i$ is in fact a set of words extracted from the action grammar for a given subtask $y_i$.  We focus on a multi-class discrimination problem where $\gamma:={1...,K}$.

A Mondrian forest classifier is constructed similarly to a random forest. Given training data $D_{1:N}$, we sample an independent collection of Mondrian Trees $T_1,...,T_M$. Each tree prediction is a distribution $p_{T_m}(y|\mathbf{x},D_{1:N})$ over the class label $y$ for a word set $\mathbf{x}$. The forest is the ensemble model averaging of each individual tree $\frac{1}{M}\sum_{m=1}^{M} p_{T_m}(y|\mathbf{x},D_{1:N})$. As the number of trees tends to infinity, the average converges to the expectation $\mathbb{E}_{T\sim MT(\lambda,D_{1:N})} [p_{T_m}(y|\mathbf{x},D_{1:N})]$. Note that the limiting expectation does not depend on the number of trees so this avoid the problem of overfitting with more sample training \cite{2014NIPS-lakshminarayanan-MondrianForests}.

For online learning, training samples are presented iteratively. At iteration $\mbox{N+1}$ each Mondrian Tree $T\sim MT(\lambda,D_{1:N})$ is updated with the new data $(\mathbf{x}_{N+1},y_{N+1})$ by sampling an extended tree $T'$ from a distribution $MTx(\lambda,T,D_{N+1})$. Using Mondrian process' properties, one can choose a probability distribution MTx such that $T'=T$ on $D_{1:N}$ is distributed according to $T\sim MT(\lambda,D_{1:N+1})$ then:
\begin{multline}\label{eqtn:onlineMondrian}
T'|T,D_{1:N+1}\sim MTx(\lambda,T,D_{N+1}) \\
\quad \mbox{implies} \quad T'\sim MT(\lambda,D_{1:N+1}).
\end{multline}
The special property of Mondrian forests is that the distribution of Mondrian trees trained on a dataset in an incremental fashion is the same as that of the latter trained in a batch fashion, regardless of the order in which data is considered. Furthermore, the complexity scales with tree depth which normally has a cost of log(N).
\section{Experiments}\label{sec:Experiments}
In this section we present the experimental setup and procedures. Three sets of experiments were conducted with , the assembly task under two coordination schemes, key LLB result detection, and an analysis of the experiments.
\subsection{Testbed Setup}\label{subsec:ExperimentalSetup}
HIRO, a 6 DoF dual-arm anthropomorph robot is driven by stiff electric actuators. The robot uses a JR3 6DoF FT sensor rigidly attached on the wrist. A specially designed end-effector tool for rigidly holding a male and female plastic camera mold was also rigidly attached to the wrist. The robot is controlled through the OpenHRP environment \cite{2004IJRR:Kaneheiro:OpenHRP}. The camera parts are designed to snap into place. In fact, the male part consists of four snap beams. A snap assembly strategy along with modular hybrid pose-force-torque controllers \cite{2013IJMA-Rojas-TwrdsSnapSensing} were used to pick up the part and then perform a set of four sub-tasks: (i) a guarded approach to the female part, (ii) a rotational alignment procedure, (iii) a snap insertion where elastic forces can be very high, and (iv) a mating procedure that maintains the parts together before moving the arm away.

\begin{figure}[t]
    \centering
        \includegraphics{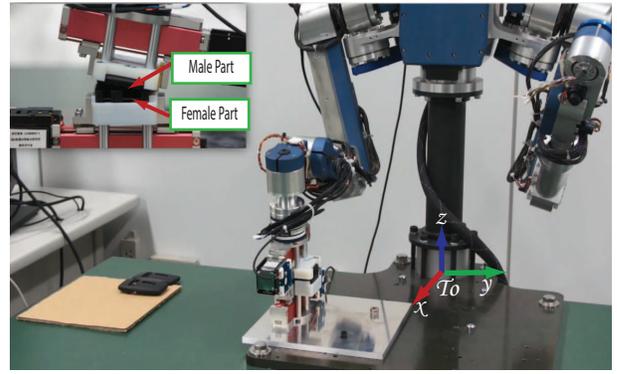}
        \caption{The HIRO humanoid robot is rigidly holding a male camera part. HIRO will run four sub-tasks to perform a snap assembly: guarded approach, a rotational alignment, a snap insertion, and a mating behavior.}
        \label{fig:ExperimentalSetup}
\end{figure}

\begin{figure}[b]
    \centering
        \includegraphics[width=3.35in, height=3in]{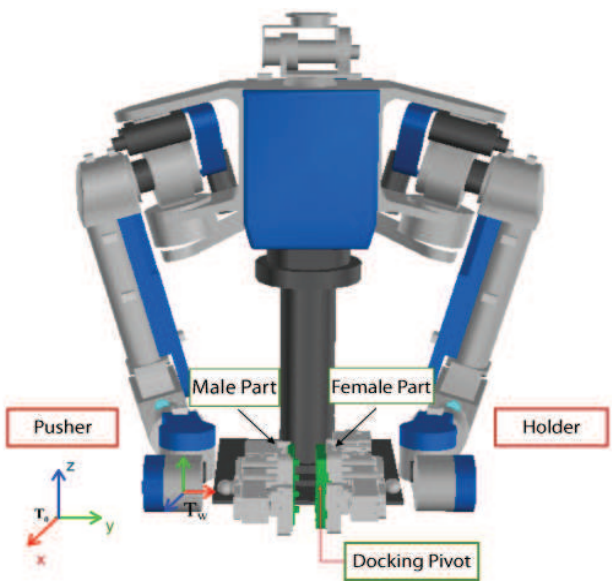}
        \caption{The HIRO dual-arm humanoid robot, rigidly holds male and female snap camera parts. Parts were assembled by having the right arm actively push while the left arm maintains a stable position under force control.}
        \label{fig:ExperimentalSetup2}
\end{figure}

The same task was performed in simulation for a one arm and a two arm scenario. We used OpenHRP's 3.0 simulation environment. Within the simulation, the male and female camera mold parts where CAD rendered from the original ones. For the two arm scenario, we tested a lateral assembly with the same strategy where the right arm functioned as the active arm and the left arm functioned as a reactionary arm with also with a force sensor and with force control to remain stead despite the push of the right arm. In one arm scenarios we segmented, encoded, and classified wrench data for only one arm. But in the two arm scenario we ran generated action grammars for both arms and performed the classification as a function of grammars in both arms. An overview of the real arm experiment setup is shown in Fig. \ref{fig:ExperimentalSetup}, whilst an illustration of the two arm simulation task is shown in Fig. \ref{fig:ExperimentalSetup2}.

The tool center point (TCP) was placed on the point in the male camera where contact with the female part would occur first for a successful task. This point served as a global reference for the system and it was provided to the system \emph{a priori}.
The world reference frame was located at the manipulator's base. The TCP position and orientation were determined with reference to the world coordinate frame $To$. The force and torque reference frames were determined with respect to the wrist's reference frame.
\subsection{Classifier Setup}\label{subsec:classifier_setup}
Classifier setup is presented first for the simulated one arm task, followed by the real one arm task, followed by the simulated two arm task.

Our goal is to classify sub-tasks to help the robot gain introspection about its action. To this end, when a success trial is used, it in fact represents an action grammar for each of the four sub-tasks of the task, for each of the six wrench axes, and for each of the 3 RCBHT levels. Fig. \ref{fig:LLBcolorCodedMap} is an illustration of the action grammar for the insertion sub-task in this assembly, for all six axes, for the LLB level. Also note that the number of words that exist for different axes in a given level and in a given state vary. To this end, we perform a resampling step that looks at the maximum number of labels for a given level and state, and extrapolate all other axis such that the same number of words exists therein.

Results are provide as the mean accuracy curve for a different number of training samples. Mean accuracy results are generated from the number of available validation trials.

For the SVM classifier we use Scikit-Learn's SVC based on libsvm for multi-class support. A radial-basis function is used for the kernel and long with an `one-versus-one' decision function shape and a penalty parameter C=1.0. As for the Mondrian Forest we used 100 trees and 12 mini batches.
\subsubsection{One Arm Simulated Robot}
The simulated one arm task consisted of 38 assembly trials, for which 4 samples are extracted that correspond to each of the 4 sub-tasks in the assembly task. Furthermore for our feature set we extract words from the action grammar for all six axes and all three levels in the RCBHT system. So for the one arm simulation training set, the feature vector consists of 1429 dimensions making the training set a (38,1429) training set.

\emph{Support Vector Machine Training}: 30 trials were used for training and 8 for validation. The classifier starts training with 1 trial (4 sub-task samples). For each successive training step, 4 new samples (corresponding to the sub-tasks) are added until 120 samples are used. The mean accuracy results are seen on row 1, column 1 of Fig. \ref{fig:results}.

\emph{Mondrian Forests}: For Mondrian Forests, the same number of training and validation is used. Training starts with 3 trials (12 samples) and 4 samples are added with each successive training step. The mean accuracy results are seen on row 1, column 2 of Fig \ref{fig:results}.

\subsubsection{One Arm Real Robot}
The real arm experiments consisted of 46 assembly trials. The same segmentation is used as with the one arm simulation. The feature length of the real arm set is the same as with the simulation: (1429,38) samples.

\emph{Support Vector Machine Training}: 36 trials were used for training and 10 for validation. The classifier starts training with 1 trial. Each successive training step adds 4 new samples until 144 samples are used. The mean accuracy results are seen on row 2, column 1 of Fig. \ref{fig:results}.

\emph{Mondrian Forests}: For Mondrian Forests, training again starts with 3 trials, 4 are added to each successive training step until 144 samples are used. The mean accuracy results are seen on row 2, column 2 of Fig \ref{fig:results}.
\subsubsection{Two Arm Simulated Robot}
The two arm simulation consisted of 20 assembly trials. The feature length for this experiment is much longer: 3786 features (1786 from the left arm and 2004 from the right arm).

\begin{figure*}[th]
        \centering
            \subfigure{\includegraphics{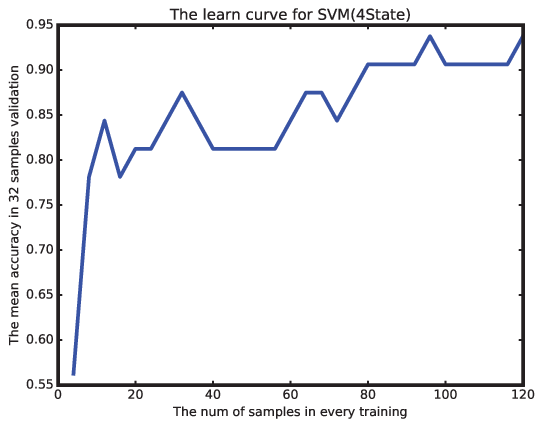}}
            \hspace{0.2cm}
            \subfigure{\includegraphics{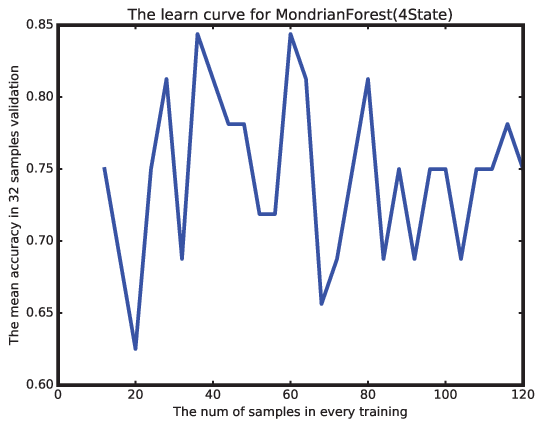}}\\
            \squeezeup
            \subfigure{\includegraphics{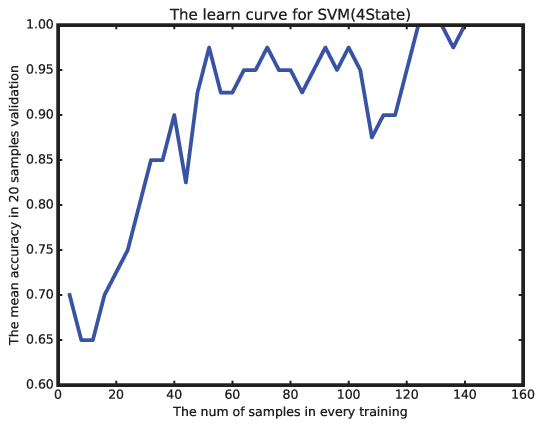}}
            \hspace{0.2cm}
            \subfigure{\includegraphics{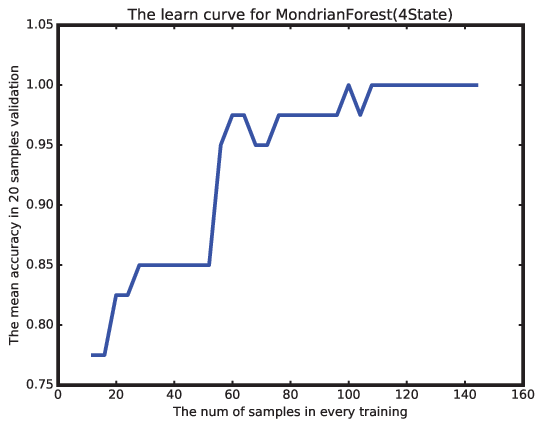}}\\
            \squeezeup
            \subfigure{\includegraphics{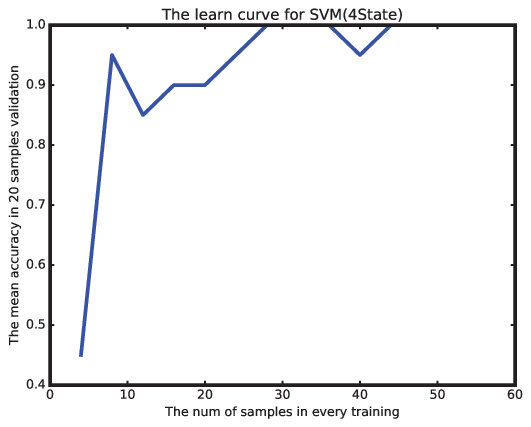}}
            \hspace{0.2cm}
            \subfigure{\includegraphics{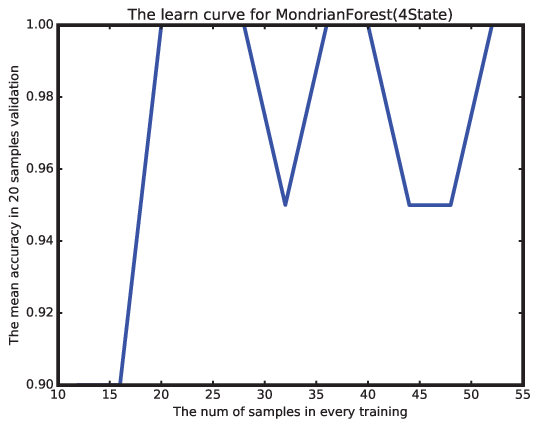}}\\

            \caption{Classification results for state introspection given a 4 state assembly task using: (i) one simulated arm, (ii) one real arm, and (iii) two simulated arms. The results correspond to the use of an SVM classifier (column 1) and a Mondrian Forests classifier (column 2).}\label{fig:results}
\end{figure*}

\emph{Support Vector Machine Training}: 14 trials were used for training and 6 for validation. The classifier starts training with 1 trial. Each successive training step adds 4 new samples until 52 samples are used. The mean accuracy results are seen on row 3, column 1 of Fig. \ref{fig:results}.

\emph{Mondrian Forests}: For Mondrian Forests, training again starts with 3 trials, 4 are added to each successive training step until 52 samples are used. The mean accuracy results are seen on row 3, column 2 of Fig \ref{fig:results}.
\subsection{Results} \label{subsec:Results}
We discuss the results of the experiments as mean accuracy values in the stead-state portion of the learning curves for all experiments and classification schemes.

We start with the real one arm experiments. In this case, the mean accuracy value for the SVM classifier was close to 95$\%$ while for Mondrian forests it was even higher, around 97.5$\%$. The two arm simulated experiment had in effect the same results: 95$\%$ accuracy using SVM and 97.5$\%$ using Mondrian forests. The one arm simulated experiment did not perform as well but still reported competitive accuracy numbers: 87$\%$ accuracy for SVM and a lower 75$\%$ for Mondrian forests.
\section{DISCUSSION} \label{sec:Discussion}
Our work shows that introspection can be bootstrapped under a segment, encode, and classify scheme. It shows that even noisy wrench signals can be encoded in this way to produce classification results with accuracies that in their steady state average at 95$\%$ or higher for both classification mechanisms for the real robot experiment and the two arm simulation and little bit lower for the one arm simulation.

The one arm simulations presented more difficulty for the classifier. The OpenHRP simulation environment does not simulate the elasticity of the snaps well, which introduces a lot of hard contacts between the snaps during the the insertion and mating portions of the assembly task. However, the classifiers were able to still separate classes with average steady state value of 87$\%$ for SVM and 75$\%$ for Mondrian forests. Perhaps the two arm simulation scenario was simpler to solve because the assembly was performed laterally instead of vertically reducing the insertion forces due to gravity and thus smoothing the insertion motion some.

Nonetheless it is well worth noting that this albeit simple scheme is able to detect robot sub-tasks in both simulated and real environments and in one arm and two arm task scenarios. In the future we wish to extend the work in a number of ways. First to test it against not only nominal sub-tasks but also with a wide variety of failure modes that results from unexpected events. Second, we want to extend the work to an online scheme that allows to perform failure correction from the unexpected scenario. Third we want to extend the work to other robots and to different tasks to further validate the generalizability of the scheme. And fourth, we are also interested in extending our scheme to a multi-modal scenario in which we use information from pose and vision, expecting that the the result will be even more reliable.

The weakness of the approach mainly lies in the manual nature of segmentation and encoding. We are currently exploring nonparametric Bayesian approaches and deep reinforcement learning approaches as well.
\section{CONCLUSION} \label{sec:Conclusion}
This work presented a principles approach to perform robot introspection of executed actions. To this end a segment-encode-and classify routine was presented including the use of an efficient online random forest algorithm and a support vector machine. The work was tested on a contact-rich snap assembly task for a one-arm assembly in simulation and with a real robot as well as a two arm simulation. The results are promising as classification accuracies were generally high across all experiments.
\section{Acknowledgments} \label{sec:Acknowledgments}
This work is supported by the grants: ``Major Project of the Guangdong Province Department for Science and Technology (2014B090919002), (2016B0911006).''
\bibliographystyle{IEEEtran}
\bibliography{\jobname}
\end{document}